\documentclass[letterpaper, 10 pt, conference]{ieeeconf}

\IEEEoverridecommandlockouts                              

\usepackage{graphics} 
\usepackage{graphicx}
\usepackage{subfigure}
\usepackage{mathtools}
\usepackage{amsmath}
\usepackage{amssymb}
\usepackage{amsthm}
\usepackage{tensor} 
\usepackage{cite}
\usepackage{bm}
\usepackage{threeparttable}
\usepackage{array}
\usepackage{multicol}  
\usepackage{multirow}
\usepackage{booktabs} 
\usepackage{verbatim}
\usepackage[utf8]{inputenc}
\usepackage{color, soul}
\usepackage[normalem]{ulem}

\theoremstyle{plain}

\soulregister\cite7
\soulregister\sout7
\soulregister\bf7

\newcommand{\Rot}{{\mathbf{R}}}

\title{\LARGE \bf IMU Data Processing For Inertial Aided Navigation: \\ A  Recurrent Neural Network Based Approach}

\author{Ming Zhang \and Mingming Zhang \and Yiming Chen \and Mingyang Li
\thanks{The authors are with Alibaba Group, Hangzhou, China. {\tt\small \{mingzhang | mingmingzhang | yimingcheng | mingyangli\}@alibaba-inc.com.}}}
\begin{document}
\maketitle
\thispagestyle{empty}
\pagestyle{empty}
\begin{abstract} 
In this work, we propose a novel method for performing inertial aided navigation, by using deep neural networks (DNNs). To date, most DNN inertial navigation methods focus on the task of inertial odometry, by taking gyroscope and accelerometer readings as input and regressing for integrated IMU poses (i.e., position and orientation). While this design has been successfully applied on a number of applications, it is {\em not} of theoretical performance guarantee unless patterned motion is involved. This inevitably leads to significantly reduced accuracy and robustness in certain use cases. 
To solve this problem, we design a framework to compute {\em observable} IMU integration terms using DNNs, followed by the numerical pose integration and sensor fusion to achieve the performance gain. 
Specifically, we perform detailed analysis on the motion terms in IMU kinematic equations, propose a dedicated network design, loss functions, and training strategies for the IMU data processing, and conduct extensive experiments.
The results show that our method is generally applicable and outperforms both traditional and DNN methods by wide margins.
\end{abstract}

\section{Introduction}

A typical inertial measurement unit (IMU) consists of a three-axis gyroscope and a three-axis accelerometer, which measure rotational velocity and gravity-compensated linear acceleration (i.e. specific force) of a moving platform respectively. 
As the IMU captures motion information without relying on any external infrastructure, it has been extensively used in navigation systems and applications, e.g., mobile devices~\cite{Li2013high,liu2020tlio}, ground robots~\cite{zhang2019localization}, miniature drones~\cite{nisar2019vimo}, autonomous vehicles~\cite{levinson2011towards}, and so on. 
Up to the hardware design and manufacturing process, the price of an IMU varies can vary over a wide range, e.g. 1-10k dollars~\cite{Goodall2019battle}. In this work, we focus on the widely-used low-cost IMUs, whose data quality is much lower than that of the high-end IMUs. This makes algorithms and systems of either improving the IMU data quality or enhancing the data processing pipeline of noteworthy significance in the research community. 

To process the IMU data, most methods exploit highly hand-crafted models to approximate sensor characteristics and the underlying motion dynamics. 
A representative pipeline can be summarized by~\cite{farrell2008aided}: data filtering (e.g., de-noising), intrinsic model compensation, data interpolation and extrapolation, pose integration, probabilistic sensor fusion or update.
The first two steps are generally recognized as data pre-processing, whose objective is to obtain a `cleaned' version of sensory input by reducing errors caused by mechanical, electronic, and signal processing imperfections. The next two steps are to calculate IMU's integrated orientation and position based on the underlying motion dynamics. To reduce the inevitable drift from the IMU integration, probabilistic update using data from other sensors~\cite{Li2013high,nisar2019vimo} or motion constraints~\cite{abdulrahim2014understanding,wagstaff2018lstm} is also implemented in a variety of applications. 

Although the \emph{state-of-the-art} IMU data processing pipeline is mature, the current design is still subject to a number of non-negligible error sources. 
One of the most critical components is the sensor modeling error introduced in the IMU data pre-processing steps, 
which has been extensively studied in the literature, especially the IMU raw data signal processing model~\cite{kang2011improvement,farrell2008aided} and the intrinsic model~\cite{trawny2005indirect,farrell2008aided,li2014high,schneider2017visual}. 
However, existing sensor models are heavily hand-crafted (e.g., number, order, and arrangement of linear and nonlinear functions in Euclidean and $SO(3)$ space), which lack of optimality guarantee and generality across different devices and practices.

In this work, inspired by recent effort of using DNN in inertial navigation algorithms~\cite{chen2018ionet,jiang2018mems,wagstaff2018lstm,yan2019ronin,brossard2020denoising}, we propose a novel data driven method for the inertial sensor modeling to achieve enhanced performance.
In particular, we design a recurrent neural network (RNN) based method to process raw gyroscope and accelerometer measurements to reduce the modeling errors and allow improved sensor fusion. 
To ensure generality of our approach under different use cases,
we train our network by regressing the observable IMU motion terms, instead of the relative IMU position and orientation that are adopted in most literature~\cite{chen2018ionet,yan2019ronin}. Additionally, we propose a dedicated method on processing IMU measurements, including multiple loss functions, training strategies, and data preparation (data pre-processing, augmentation, and so on),  which are critical in 
applying our method for high-precision inertial aided navigation.
To validate the effectiveness of the proposed method, extensive real-world experiments are also conducted in this work, by utilizing the datasets collected from automotive vehicles~\cite{choi2018kaist}, drones~\cite{burri2016euroc}, and ground robots. The results show that the proposed method outperforms the various competing methods, including both traditional and DNN based ones~\cite{farrell2008aided,brossard2020denoising,liu2020tlio,silva2019end,zhang2019localization}, by wide margins.
\section{Related Work}
Low-cost IMUs are subject to multiple error sources, including the ones coming from mechanical, electronic, and signal processing imperfections.
To date, most existing methods on reducing the noises in IMU readings and compensating for the intrinsic models are based on statistical approaches. Standard methods in filtering the IMU readings in inertial navigation are to eliminate or attenuate the high frequency components~\cite{kang2011improvement}. Auto regressing and moving average (ARMA) methods are also widely applied in this field~\cite{quinchia2013comparison}. In addition to noises, IMU readings are also affected by sensor biases, which are generally modeled as additive components in the measurement equations~\cite{trawny2005indirect}. For high-end IMUs, modeling noises and biases might be enough for high-precision navigation~\cite{farrell2008aided}. However, to have high-quality performance, additional parameters must be considered for low-cost IMUs, including scale factors, misalignment parameters, G-sensitivity matrix, and other high order terms~\cite{li2014high,yang2020online}. Compensating for the thermal effects is another important factor that is widely studied and must be taken into consideration in real applications~\cite{niu2013fast}.
Once readings from IMUs are properly filtered and a closed-form sensor model equation is given, IMU poses can be straightforwardly computed by numerical integration following the Newton's law~\cite{farrell2008aided,Li2013high}.

However, due to the hardware complexity of low-cost IMUs, it becomes almost infeasible for researchers and engineers to formulate the {\em exact} mathematical equations to describe the sensor noise and intrinsic models. Therefore, using data driven methods instead of the model based ones in this domain becomes worthy exploring. Motivated by the recent success of DNN in a variety of tasks, researchers started to look into methods that represent IMU models with DNNs to improve the inertial navigation performance~\cite{chen2018ionet,wagstaff2018lstm,yan2019ronin,jiang2018mems,brossard2020denoising,liu2020tlio}. 
The majority of existing DNN methods are designed to perform the end-to-end IMU integration~\cite{chen2018ionet,wagstaff2018lstm,yan2019ronin}, by taking sensor data as inputs and integrating poses using neural networks. 
However, such design regresses both observable or un-observable system terms~\cite{li2014high}, so its robustness can only be guaranteed when patterned motions are involved. Alternatively, there are DNN based methods that target on refining sensor models~\cite{jiang2018mems,brossard2020denoising}.   
However, both~\cite{jiang2018mems,brossard2020denoising} can not be applied to general-case pose integration since they are either optimized for stationary periods or only able to handle rotations.  

Since noises and biases in IMU measurements are unavoidable, even with proper sensor model, the accumulated drift in the integrated IMU poses will inevitably occur. To reduce the long-term pose drift, there are generally two families of algorithms: by fusing IMU readings with
measurements from other sensors~\cite{farrell2008aided,Li2013high,geneva2018lips,zhang2019localization} or relying on motion constraints of the platform where the IMU is installed on~\cite{abdulrahim2014understanding,wagstaff2018lstm,ahmed2018visual}.
Representative sensors that are widely used in combination with IMU include global positioning system (GPS)~\cite{farrell2008aided}, visual sensors~\cite{Li2013high}, laser range finders~\cite{geneva2018lips}, wheel odometers~\cite{zhang2019localization}, and so on. On the other hand, if repetitive or special motion patterns are involved in the inertial navigation, they can also be probabilistically utilized to reduce the estimation errors. Typical motion patterns used in the navigation systems are zero velocity events~\cite{abdulrahim2014understanding,wagstaff2018lstm} or pedestrian patterns~\cite{ahmed2018visual}.
There are also algorithms that utilize both measurements from complementary sensors and motion constraints to reduce pose errors~\cite{kottas2013detecting,nisar2019vimo}. For instance, Kottas et al.~\cite{kottas2013detecting} proposed a method on visual-inertial odometry by integrating a module of hovering motion detection, and VIMO~\cite{nisar2019vimo} designed a system that simultaneously estimates IMU poses and force using multi-sensory inputs. 

In this work, we focus on proposing a DNN based method that is to model IMU measurements using data driven methodology. In addition, we also ensure that the filtered IMU readings can be fused with measurements from other sensors to conduct inertial navigation, with higher accuracy compared to competing methods. The rest of the paper is organized as follows. Section~\ref{sec:model} provides mathematical details on IMU integration equations, discusses on generally learn-able terms, and presents our design choices. Section~\ref{sec:learning} describes the details of our DNN design, including the network architecture, loss functions, training strategies, and data preparation. Finally, Section~\ref{sec:result} presents our experimental results from multiple testing platforms which validate the advantages of our method.  
\section{IMU models}
\label{sec:model}
\subsection{IMU Measurement Equations}
In this work, we assume an IMU, $\{I\}$, moves with respect to a global frame, $\{G\}$. The corresponding IMU measurement model can be described by:
\begin{align}
\label{eqn:imu_general} 
[{\bm \omega}^T_{rm}, {\bm a}^T_{rm}]^T = 
f_{\pi} \left({^I\bm \omega},{^I}{\bm a}, {\bm \pi}, {\bm n} \right)
\end{align}
where ${\bm \omega}_{rm}$ and ${\bm a}_{rm}$ are the \textbf{raw} gyroscope and accelerometer data,
${^I}{\bm \omega}$ the angular rate in the IMU frame $\{I\}$, $f_{\pi}$ the IMU measurement model and $\bm \pi$ the model parameter vector, and $\bm n$ the noise terms. Additionally,
${^I}{\bm a} = {^I_G}{\bf R}({^G}{\bm a} \!-\! {^G}{\bm g})$, where $^G\bm a$ is the linear acceleration in global frame $\{G\}$, ${^I_G}{\bf R}$ the rotation from $\{G\}$ to $\{I\}$, and ${^G}{\bm g}$ the known gravity (e.g., ${^G}{\bm g} = [0,0,-9.8m/s^2]^T$).

To design $f_\pi$, most inertial navigation algorithms rely on hand-crafted pre-processing steps for measurement de-noising and intrinsic compensation:
\begin{align}
\label{eqn:imu_pre} 
[{\bm \omega}^T_{m}, {\bm a}^T_{m}]^T = 
g_{\pi} \left({\bm \omega_{rm}},{\bm a_{rm}},{\bm \pi} \right)
\end{align}
where $g_{\pi}$ is the {pre-processing} model, 
${\bm \omega}_{m}$ and ${\bm a}_{m}$ the {pre-processed} measurements, commonly approximated by 
linear models, e.g., bias model:~\cite{farrell2008aided,trawny2005indirect,Li2013high}:
\begin{align}
\label{eqn:w_m} 
{\bm \omega}_m \!=\! {^I}{\bm \omega} \!+\! {\bm b}_g \!+\! {\bm n}_g,\,\,\,
{\bm a}_m \!=\!  {^I}{\bm a} \!+ \!{\bm b}_a \!+\! {\bm n}_a
\end{align}
where ${\bm b}_g$ and ${\bm b}_a$ the biases and ${\bm n}_g$ and ${\bm n}_a$ measurement noises.
Our objective is to represent $g_\pi$ and $\bm \pi$ via a data driven way to achieve better performance compared to 
existing hand-crafted methods.

\subsection{IMU Based Pose Integration}
To describe the detailed formulation of our method, we first present the IMU state vector and integration equations that are generally used in the inertial navigation. 
Following~\cite{trawny2005indirect,Li2013high,schneider2018maplab}, the IMU state is defined as:
\begin{align}
\label{eq:state_single}
{\bm x} =
\left[
{^{G}_{I}{\bar{\bm q}}}^\intercal, ~
{\bm b}^\intercal_{g}, ~
{{^G}{\bm v}_I}^\intercal, ~
{\bm b}^\intercal_{a}, ~
{{^G}{\bm p}_I}^\intercal
~\right]^\intercal \in \mathbb{R}^{16}
\end{align}
where ${^{G}_{I}{\bar{\bm q}}}$ presents rotation from IMU frame to global frame (i.e. ${^G_I}\Rot$) in quaternion~\cite{trawny2005indirect}, and ${{^G}{\bm p}_I}$ and ${{^G}{\bm v}}_I$ stand for the IMU's position and velocity. 


Denoting ${^{G}_{I}{\bar{\bm q}}}(t_k)$, ${{^G}{\bm p}_I} (t_k)$, and ${{^G}{\bm v}}_I (t_k)$ for IMU's rotation, position, and velocity 
at time $t_k$, the IMU integration kinematic equations are: 
\begin{align}
\label{eq:rotint}
{^{G}_{I}{\bar{\bm q}}}(t_{k+1}) &=
{^{G}_{I}{\bar{\bm q}}}(t_{k}) 
{}
\Delta{{\bar{\bm q}}} \\
\label{eq:velint}
^{G} {\bm v}_{I}(t_{k+1}) &= {}
^{G} {\bm v}_{I}(t_{k}) +
\int _{t_k}^{t_{k+1}} 
{^G \bm a}_{I}(\tau) d \tau \notag \\
&= {} ^{G} {\bm v}_{I}(t_{k}) +
{^G}{\bm g} \Delta t + 
{^G_I \bm R}(t_k)
{{^{I(t_{k+1})}_{I(t_{k})}{\bm \beta}}} \\
^{G} {\bm p}_{I}(t_{k+1}) &= {}
^{G} {\bm p}_{I}(t_{k})  {} +
^{G} {\bm v}_{I}(t_{k}) \Delta t \!+\!\!
\int _{t_k}^{t_{k+1}} \!\!\! 
\int _{t_k}^{\tau} \!\!
{^G \bm a}_{I}(s) d s d \tau \notag \\
 &={} ^{G} {\bm p}_{I}(t_{k}) +{}
^{G} {\bm v}_{I}(t_{k}) \Delta t +
\frac{1}{2} {^G}{\bm g} (\Delta t)^2 \notag \\ 
\label{eq:posint}
&\,\,\,\,\,\,\,\,\,\,\,+
{^G_I \bm R}(t_k) {{^{I(t_{k+1})}_{I(t_{k})}{\bm \gamma}}}
\end{align}
where , 
$\Delta t = t_{k+1} - t_{k}$, and 
\begin{align}
\label{temp1}
{}_{I(t_{k})}^{I(t_{k+1})}{{\bm \beta}} &= 
\int _{t_k}^{t_{k+1}} 
{^{I(t_k)}_{I(\tau)} \bm R} \cdot {^{I(\tau)} \bm a} d \tau \\
\label{temp2}
{}_{I(t_{k})}^{I(t_{k+1})}{{\bm \gamma}} &= 
\int _{t_k}^{t_{k+1}} 
\int _{t_k}^{\tau} 
{^{I(t_k)}_{I(s)} \bm R} \cdot {^{I(s)} \bm a} d s d \tau 
\end{align} 
We also define $\Delta{{\bar{\bm q}}} = {}_{I(t_{k})}^{I(t_{k+1})}{\bar{\bm q}}$, 
$\Delta{\bm \beta} = {}_{I(t_{k})}^{I(t_{k+1})}{{\bm \beta}}$,
$\Delta{{{\bm \gamma}}} = {}_{I(t_{k})}^{I(t_{k+1})}{{\bm \gamma}}$ as simplified notation for later usage in this work.
We note that in the above equations $\Delta {\bar{\bm q}}$, 
$\Delta {\bm \beta}$,
and $\Delta {\bm \gamma}$ are fully characterized by $^{I(t)}\bm \omega$ and $^{I(t)}\bm a$, $t_k\leq t \leq t_{k+1}$, and independent of all other variables. All intermediate rotation terms in Eqs.~\ref{temp1}
and~\ref{temp2}, i.e. ${^{I(t_k)}_{I(\tau)} \bm R}$ and 
${^{I(t_k)}_{I(s)} \bm R}$, can be computed by integrating $^{I(t)}\bm \omega$.

\subsection{Learn-able Terms}
Existing DNN methods seek to directly regress the relative position and rotation terms~\cite{chen2018ionet,silva2019end,yan2019ronin}:
\begin{align}
\label{eq:delta q}
\Delta {\bm q} &\!=\! 
{^{G}_{I}{\bar{\bm q}}}(t_{k})^{-1} 
{^{G}_{I}{\bar{\bm q}}}(t_{k+1}) \\
\Delta {\bm p} &\!=\! 
{^G_I \bm R^T}(t_k) \left( ^{G} {\bm p}_{I}(t_{k+1}) \!-\! {}
^{G} {\bm p}_{I}(t_{k}) \right)
\end{align} 
While computing $\Delta {\bar{\bm q}}$ is feasible via
Eq.~\ref{eq:rotint}, obtaining the  relative position term 
$\Delta {\bm p}$ requires $^{I(t_k)} {\bm v}_{I}(t_{k})$ and ${^{I(t_k)}}{\bm g} = {^G_I \bm R^T}(t_k) {^G\bm g}$ from Eq.~\ref{eq:posint}.
If a DNN method is able to compute $\Delta {\bm p}$ by using windowed IMU measurements, this network must also obtain $^{I(t_k)} {\bm v}_{I}(t_{k})$ and ${^{I(t_k)}}{\bm g}$ as hidden variables. However, both terms are {\em not} necessarily observable using IMU measurements~\cite{farrell2008aided}, unless patterned motion~\cite{chen2018ionet}, zero velocity event~\cite{wagstaff2018lstm}, or other special cases are involved. In other words, it is always possible that under the same IMU measurements the corresponding position ($\Delta {\bm p}$) is different.

To avoid this problem and allow general usage, we propose to leverage RNN methods to compute the learnable terms $\Delta {\bm \gamma}$, $\Delta {\bm \beta}$, and $\Delta {\bm q}$ instead of the relative poses. As analyzed previously, all those terms are functions of the motion dynamic terms $^{I}\bm \omega$ and $^{I}\bm a$ only, and independent of prior information on local gravity direction, local velocity term, and so on. Since $^{I}\bm \omega$ and $^{I}\bm a$ are both measured by the IMU sensor, our method directly captures the intrinsic model of an IMU. By contrast, the network in previous methods~\cite{chen2018ionet,silva2019end,yan2019ronin} express both IMU models and motion patterns in the training datasets.
Our design also aligns well with the closed-form solutions that 
use $\Delta {\bm \gamma}$, $\Delta {\bm \beta}$, and $\Delta {\bm q}$ for later sensor fusion~\cite{Li2013high, forster2017manifold}.

\begin{figure*}
	\centering
	\includegraphics[width=0.8\linewidth]{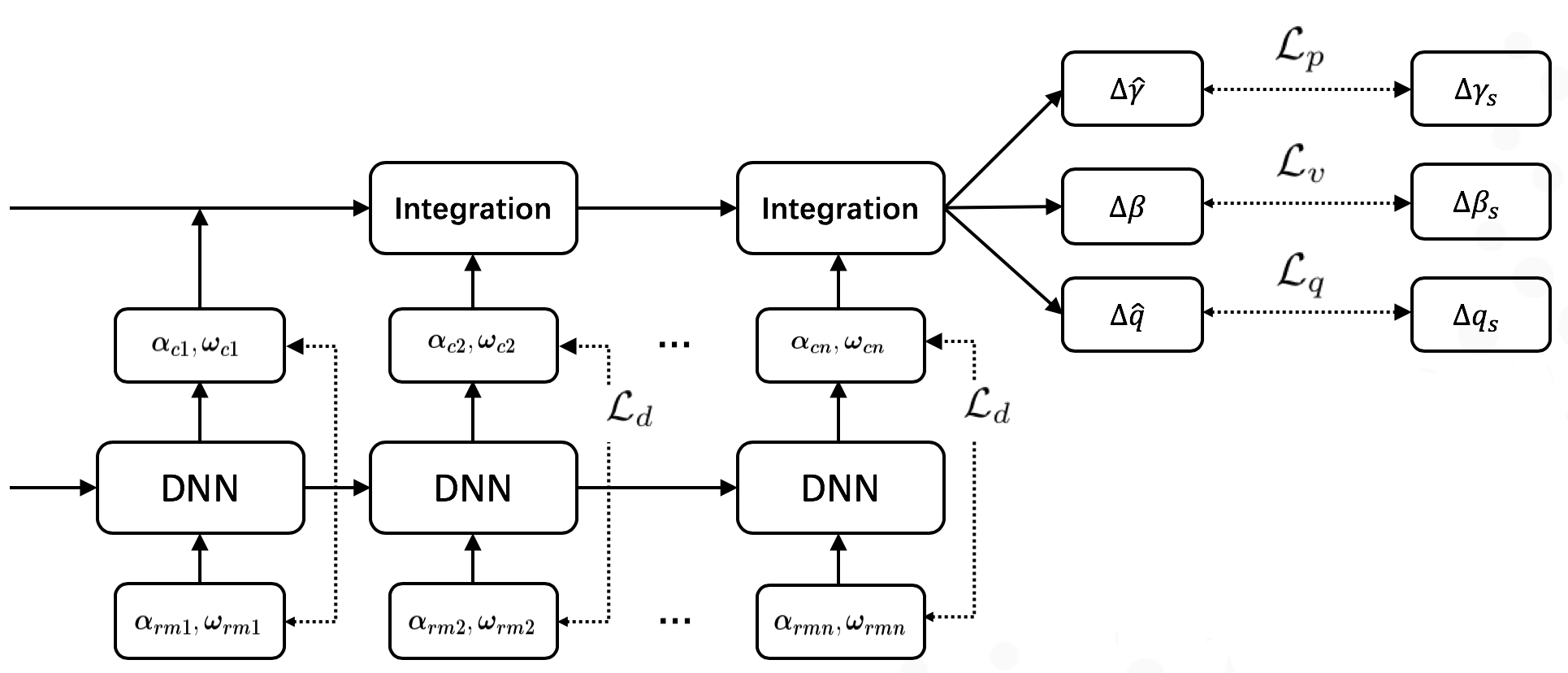}
	\caption{The proposed network and learnable terms.}
	\label{fig:pipeline}
\end{figure*}

\section{Learning Method}
\label{sec:learning}
\subsection{Network Architecture}
The overall architecture of our network is shown in Fig.~\ref{fig:pipeline}, which consists of both standard RNN and closed-form IMU integration modules. 
Specifically, the RNN module is used to represent the IMU model (de-noise, intrinsic, and others), 
which takes raw measurements as input and calculates refined values.
The differentiable integration module is used to generate $\Delta {\bm \gamma}, \Delta {\bm \beta}$ and $\Delta {\bm q}$ for loss functions.  
We note that, an alternative solution is to use RNN layers to directly regress $\Delta {\bm \gamma}, \Delta {\bm \beta}$ and $\Delta {\bm q}$ instead of adding the extra integration blocks. However, integration computation requires a combination of operations in 
Euclidean and $SO(3)$ space (especially the ones relate to rotation terms), which can not be well expressed by standard light-weighted RNN layers. 
In our method,
we use stacked bidirectional LSTM connected by fully-connected (FC) layers for RNN, and utilize the parameter-free closed-form differential equations~\cite{Li2013high} for the integration module. 

In the training stage, the output of RNN layer will be passed to the integration layer for calculating the training loss. In the reference stage, the RNN layer computes the learned IMU measurements, which will be directly used as the input for the probabilistic sensor fusion. 
We note that, our network architecture is relatively preliminary, which can be further largely optimized via a number of methods (e.g., using~\cite{vaswani2017attention}). 
However, our results show that the current design already outperforms competing methods significantly, which strongly support our design purpose of using DNNs to process IMU data in inertial navigation.

\subsection{Loss Function}
For the task of learning high precision IMU models, the ideal method is to design a loss function to measure the error between the learned IMU measurements and the ground truth ones. However, such design is not practical since it is almost impossible to obtain the `ground-truth' rotational velocities and linear accelerations of real IMU hardware devices. Therefore, we must utilize the indirect loss terms on the training stage. In this work, we define loss terms as:
\begin{align}
\label{eq:indirect_loss}
\mathcal{L}_{q} &=
|
log(\Delta {\bm q}_{s} \otimes \Delta \hat{\bm q}^{-1})|_{h} \\
\label{eq:indirect_loss2}
\mathcal{L}_{v} &=
|\Delta {\bm \beta}_{s} - \Delta \hat{\bm \beta}|_{h} \\
\label{eq:indirect_loss3}
\mathcal{L}_{p} &=
|\Delta {\bm \gamma}_{s} - \Delta \hat{\bm \gamma}|_{h}
\end{align}
where ${\bm a}_s$ and $\hat{\bm{a}}$ represent ground-truth and DNN output value of the variable ${\bm a}$, $|\cdot|_h$ denotes the Huber loss function, $\otimes$ is the quaternion multiplication operator, and 
$log(\cdot)$ is the logarithm operator in $SO(3)$ space.
It is important to note that, in this work, the ground-truth used in training are provided by either additional high-quality sensors or sensor fusion algorithms (see Sec.~\ref{sec:result} for details). Therefore, the selected Huber loss operator is able to improve training robustness, to systematically handle possible errors and outliers in the training ground truth.

Additionally, we use a regularization loss 
\begin{align}
\label{eq:direct_loss}
\mathcal{L}_{d} = 
max \left(
|{\bm u}_m - \hat{\bm u}| - \lambda,0
\right)
\end{align}
where ${\bm u}_m$ is the raw IMU measurement, $\hat{\bm u}$ is the refined one, and $\lambda$ is a control parameter. This term penalizes the refined IMU values only when they deviate from the raw measurements above a threshold, which is to ensure fast network convergence. 
The design motivation of this term is as follows. 
Although the IMU measurements are subject to multiple error sources, directly integrating IMU data will still lead to poses of certain quality. To train a network, we would like to start from that `certain' quality, instead of starting from scratch. Additionally, this term also ensures that the IMU intrinsic models only incur a certain amount of changes in the corresponding measurements values and reject unlimited changes, which stem from the process of how the IMU measurements are generated electronically.  

\subsection{Training}
The ground-truth training signals $\Delta {\bm q}_{s}$, $\Delta {\bm \beta}_{s}$, and $\Delta {\bm \gamma}_{s}$, as shown later in experiments, are derived through manipulating Eq.~\ref{eq:velint} and~\ref{eq:posint} with the ground-truth poses. The ground-truth poses are from either high-fidelity sensors or external infrastructures.
Additionally,  Fig.~\ref{fig:pipeline} shows that we compute the loss function by integrating all measurements within a time window. Furthermore, to enhance the training, integration is conducted by integrating $20\%$, $40\%$, $60\%$, $80\%$, and $100\%$ data to formulate multiple loss terms. This is designed since sensor fusion algorithm using IMU measurements might need integration for different duration times and this is naturally the `data augmentation' for better training results.

To enhance our network performance, we also propose dedicated data augmentation strategies. Specifically, we start with different indexes of IMU data to make windowed IMU input for training, and add manually generated noise into raw IMU measurements. Furthermore, since our refined measurement still contains bias~\footnote{Using the trained model to de-bias completely at the inference stage is not practically feasible, since the bias is time-varying.}, we also manually add random biases during training. The loss function thus becomes:
\begin{align}
\mathcal{L}_{q} &\!=\!
|
log(\Delta {\bm q}_{s} \!\otimes\! \Delta \hat{\bm q}^{-1} \!\otimes\! {\bm q}_b^{-1})|_{h} \\
\mathcal{L}_{v} &\!=\!
|\Delta {\bm \beta}_{s} \!-\! \Delta \hat{\bm \beta} \!-\! \hat{\bm \beta}_b|_{h} \\
\mathcal{L}_{p} &\!=\!
|\Delta {\bm \gamma}_{s} - \Delta \hat{\bm \gamma}\! -\! \hat{\bm \gamma}_b|_{h} \notag
\end{align}
where $\hat{\bm q}_b$, $\hat{\bm \beta}_b$,  and $\hat{\bm \gamma}_b$ are the integrated biases.

\section{Experiments and Results}
\label{sec:result}
\subsection{Setup}

\begin{table*} [t]
	\centering
	\begin{threeparttable}  
		\caption{\scriptsize{Positional RMSE (m) over the entire trajectories, using the five test sequences in the EuRoC dataset..}}  \label{table: trajectory_rmse}
		\begin{tabular}{l p{2.0cm}<{\centering}
				p{2.5cm}<{\centering}
				p{2.5cm}<{\centering}
				p{2.5cm}<{\centering}
				p{2.5cm}<{\centering}
				p{2.5cm}<{\centering}
			}  
			
			\toprule  
			\quad Sequence & Tr.~\cite{farrell2008aided} & VINS~\cite{qin2018vins} & TLIO~\cite{liu2020tlio} & End-to-End learning~\cite{silva2019end} & Pr. \\
			\midrule
			\quad MH\_02\_easy  & 0.185 & 0.150 & 0.303 &  3.307 & {\bf 0.150} \\
			\quad MH\_04\_difficult & 0.151 & 0.320 & 0.823 &  5.199 & {\bf 0.138} \\
			\quad V1\_03\_difficult & 0.267 &0.180  & 0.653 &  5.329 & {\bf 0.147} \\
			\quad V2\_02\_medium & 0.111 & 0.160 & 0.509 &  2.897 & {\bf 0.104} \\
			\quad V1\_01\_easy & 0.081 & 0.079 & 0.343 &  8.166 & {\bf 0.066} \\
			\quad Average & 0.159 & 0.179 & 0.526 &  4.980 & {\bf 0.121} \\
			\bottomrule  
		\end{tabular}  
	\end{threeparttable}  
\end{table*} 
\begin{table*}
	\centering
	\begin{threeparttable}
		\caption{\scriptsize{RMSE in relative translation (m) and orientation (rad) over ten IMU frames for different methods, using the five test sequences in the EuRoC dataset. The reported numbers are in the format of (translation (m) / orientation (rad)).}}
		\label{table:delta_pose_error}
		\begin{tabular}{l p{2.0cm}<{\centering}
				p{2.5cm}<{\centering}
				p{2.5cm}<{\centering}
				p{2.8cm}<{\centering}
				p{2.2cm}<{\centering}
			}
			\toprule
			\quad Sequence & Tr.~\cite{farrell2008aided} & Gyro-Denoising~\cite{brossard2020denoising} & TLIO~\cite{liu2020tlio} & End-to-End learning~\cite{silva2019end} & Pr.\\
			\midrule
			\quad MH\_02\_easy & 0.0108 / 0.0093 & - / 0.0060 & 0.0053 / 0.0045 &  0.0077 / 0.0072 & {\bf 0.0028} / {\bf 0.0025} \\
			\quad MH\_04\_difficult & 0.0237 / 0.0103 & - / 0.0061 & 0.0078 / 0.0058 & 0.0092 / 0.0078 & {\bf 0.0058} / {\bf 0.0037} \\
			\quad V1\_03\_difficult & 0.0189 / 0.0168 & - / {\bf 0.0051} & 0.0151 / 0.0099 & 0.0130 / 0.0107 & {\bf 0.0053} / 0.0067 \\
			\quad V2\_02\_medium & 0.0191 / 0.0130 & - / 0.0118 & 0.0127 / 0.0083 & 0.0131 / 0.0122 & {\bf 0.0083} / {\bf 0.0068} \\
			\quad V1\_01\_easy & 0.0118 / 0.0234 & - / 0.0168 & 0.0053 / 0.0181 & 0.0075 / 0.0219 & {\bf 0.0030} / {\bf 0.0162}  \\
			\quad Average & 0.0169 / 0.0146 & - / 0.0092 & 0.0092 / 0.0093 & 0.0101 / 0.0120 & {\bf 0.0050} / {\bf 0.0072}  \\
			\bottomrule
		\end{tabular}
	\end{threeparttable}
\end{table*}

\begin{figure*}[t]
	\centering
	\includegraphics[scale=0.35]{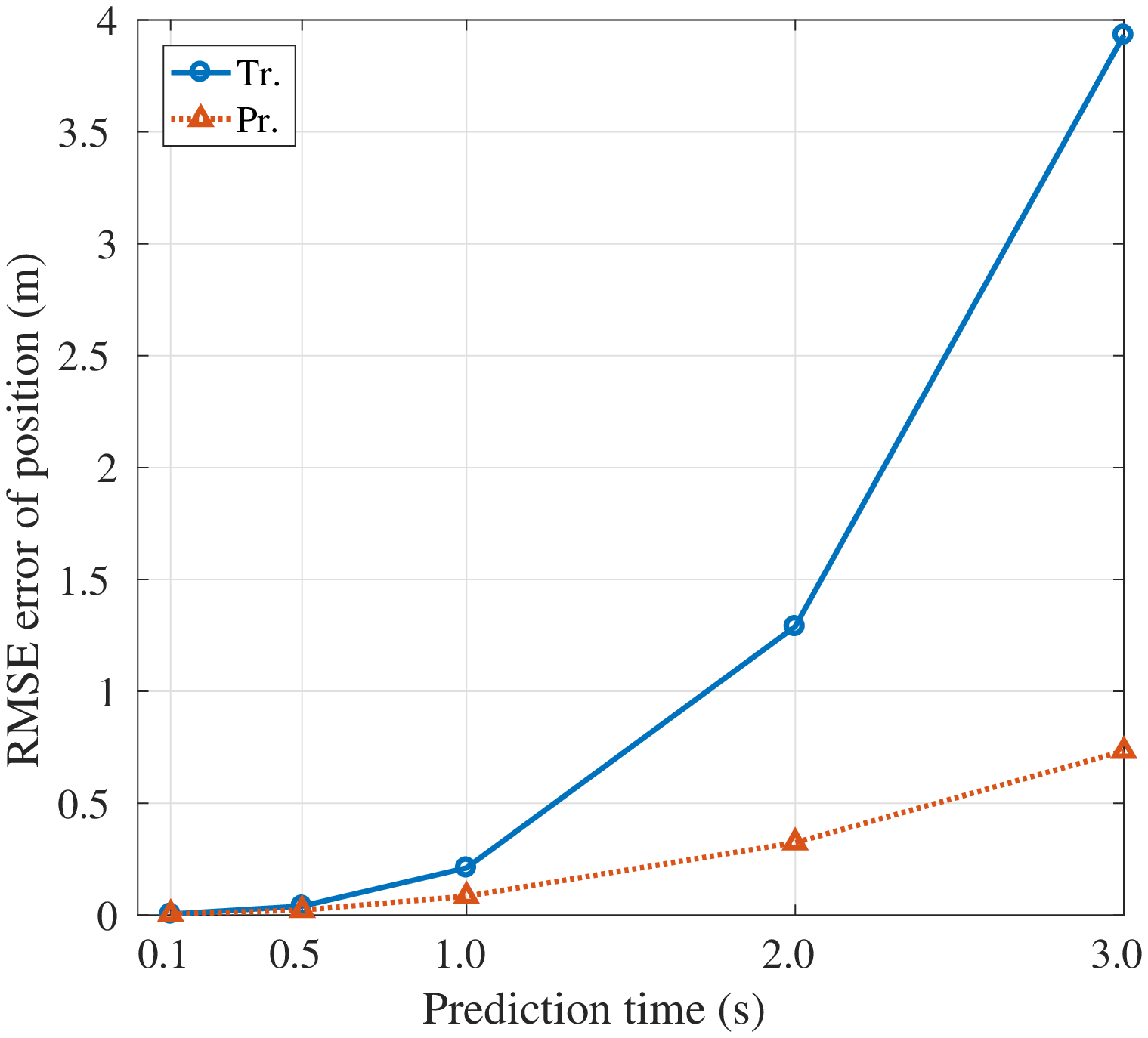}
	\includegraphics[scale=0.35]{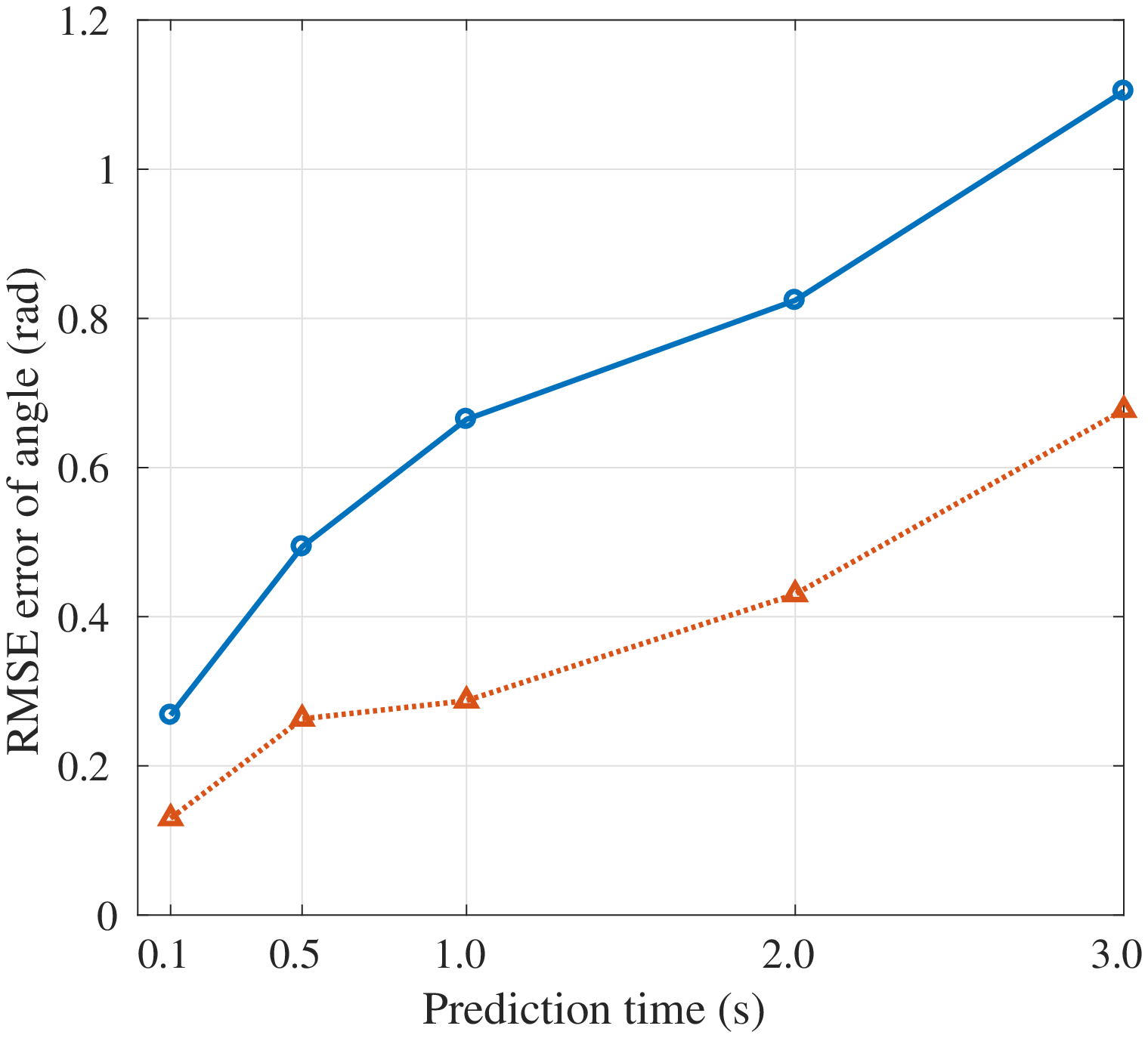}
	\includegraphics[scale=0.35]{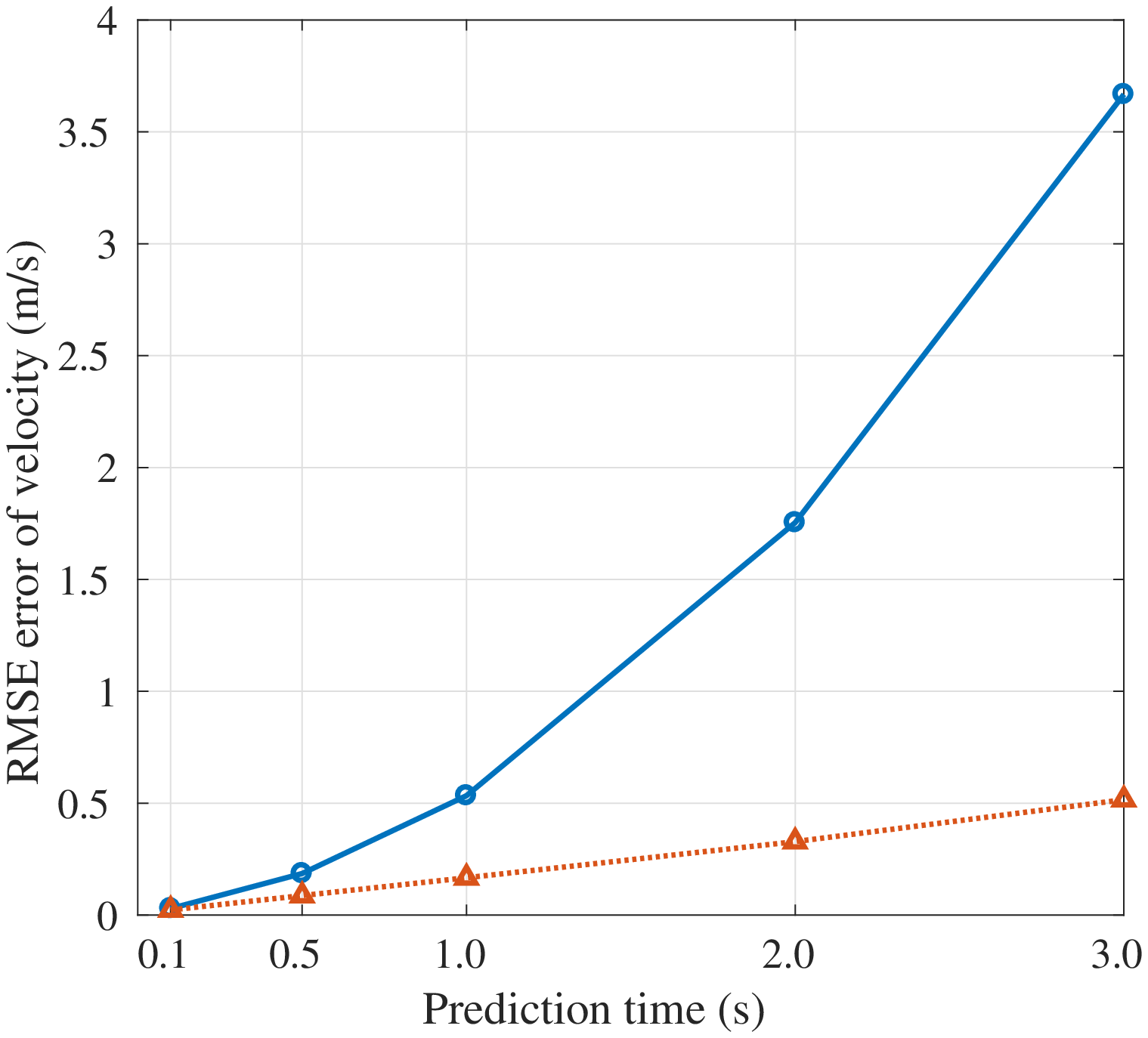}
	\caption{IMU integration errors for varying time periods, using the traditional method and proposed one. Left: Errors in predicted position; Middle: Errors in predicted orientation; Right: Errors in predicted velocity.}
	\label{fig:prediction_error}
\end{figure*}

To demonstrate the performance of the proposed method, we conducted experiments on three platforms: 
\begin{enumerate}
    \item EuRoC~\cite{burri2016euroc}: the dataset contains image and inertial data from a drone. The measurements recorded by the image sensors and IMU were 20Hz and 200Hz respectively. 
    Poses provided from a Vicon motion capture system are used as ground truth for training and also for computing RMS errors for testing.
    \item KAIST~\cite{choi2018kaist}: collected from automotive vehicles in urban scenes. A sensor fusion method using RTK-GPS, an IMU, and 3D LiDAR is used to generate the ground truth poses.
    The measurements recorded by the image sensors and IMU were 10Hz and 100Hz respectively.
    \item Ground robots: collected from a ground robot (Clearpath), with 10Hz images and 200Hz IMU data. Ground truth poses are computed by RTK-GPS assisted sensor fusion algorithm (RTK-VIO). Since the RTK-VIO algorithm outputs high precision poses as the image is received, the ground truth pose is not at the exact IMU measurement timestamp. We interpolate the ground truth pose to align with the nearest IMU timestamp.
\end{enumerate}
For each dataset, we randomly chose half of the sequences for training and the others for testing. 
Our method was implemented on Tensorflow 2.2.0, in which ADAM optimizer was used with a learning rate initialized at 0.0001 and training epochs at 700. The training was carried out on a single NVIDIA GeForce GTX 1080 Ti GPU with a batch size of 32 samples. The model with the best validation loss throughout the training was chosen as the final one for the testing.

\subsection{Competing Methods}
We compared the proposed method (termed Pr.) against five competing algorithms:
\begin{enumerate}
    \item Traditional method (termed Tr.): This method is to pre-process IMU data using the traditional approaches, i.e., a low-pass filter followed by the intrinsic modeling~\cite{farrell2008aided}. The results are used either standalone for integration or fused with the visual data for long term navigation~\cite{zhang2019localization}. 
    \item VINS~\cite{qin2018vins}: The second one is a representative state-of-the-art visual-inertial odometry method with open-source implementation, which also relies on analytical equations to process IMU measurements.
    \item End-to-end learning (termed EL.) method~\cite{silva2019end}: The third one is a competing DNN method by learning the relative position and rotation directly, with our customized RNN implementation. 
    \item TLIO~\cite{liu2020tlio}: The fourth method is the EKF and deep learning combined method, which demonstrates high-quality performance in augmented reality applications. 
    \item Gyro de-noising~\cite{brossard2020denoising}: The last one is a DNN based gyroscope de-noising method for attitude estimation.
\end{enumerate}
We note that the formulation of the first two competing methods and the proposed one allow for sensor fusion with visual sensors, while other methods yield pose estimates only based on IMU measurements. Additionally, the gyro de-noising~\cite{brossard2020denoising} method only computes rotational estimates, while all other methods compute both positional and rotational measurements. The motivation of also comparing~\cite{brossard2020denoising} in experiments is due to the fact that the algorithm design in that work is similar to us by pre-filtering the sensory data. 

\subsection{Tests on EuRoC Dataset}
Since EuRoC~\cite{burri2016euroc} contains high-quality ground truth values for both rotational and positional terms, more detailed tests are performed under the corresponding sequences.

\textbf{Long-term Inertial Navigation Accuracy Test:}
The first test is to demonstrate the most important results: the pose estimation accuracy.
In this test, the processed IMU measurements by both the proposed and the traditional methods are fused respectively with only visual sensor data using the approach of~\cite{zhang2019localization}, 
for open-loop positional estimation. 
On the other hand, the EL.~\cite{silva2019end} and TLIO~\cite{liu2020tlio} methods only utilize IMU measurements. 
Note that, in this test, we sought to obtain the `best possible' inertial navigation accuracy for all competing algorithms. By computing observable IMU terms, both Tr. and Pr. allow sensor fusion probabilistically. On the other hand, since EL.~\cite{silva2019end} and TLIO~\cite{liu2020tlio} directly regress for IMU's relative poses, the standard methods of fusing visual and inertial data become not applicable. 
Results of VINS are the reported ones in~\cite{qin2018vins}.


In this test, we split the training and testing sequences similar to the tests conducted in the original paper of EL.~\cite{silva2019end}. To show the results,
we compared the root-mean-squared errors (RMSE) between the ground-truth IMU positions and the estimated ones by different algorithms. 
The results are shown in Table~\ref{table: trajectory_rmse}. Those computed positional RMSE values clearly demonstrate that the proposed method outperforms competing ones by wide margins.
Compared to traditional method and VINS~\cite{qin2018vins}, our method is able to better represent the IMU model. Compared to the DNN methods, our method regresses learn-able terms which are of better accuracy and make probabilistic sensor fusion possible. 

\textbf{Relative Pose Accuracy:}
In addition to knowing the long-term inertial navigation accuracy, it is also interesting to investigate the relative pose accuracy within short time windows. In this test, the RMSE for the relative position and orientation over 10 IMU readings are computed for different algorithms, and the results are reported in Table.~\ref{table:delta_pose_error}. 
We note that, compared to the previous test, we also added~\cite{brossard2020denoising} into the competing algorithms, which is an algorithm that only seeks to regress for rotational estimates. 
We also note that visual measurements are not used in Tr. and Pr. and VINS~\cite{qin2018vins} is not compared against here, due to the goal of looking into the short-term pose accuracy.

Results in Table.~\ref{table:delta_pose_error} show that the proposed method outperforms all competing methods clearly. The average RMSE of position and orientation of our method are 54\% and 78\% lower than the best competing method. This test clearly demonstrates the advantage of our method, including both loss function design and the training strategies.

\textbf{Pose Integration Accuracy:} 
The next test is to demonstrate the accuracy of IMU prediction between varying time periods for the proposed and traditional methods. This is due to the fact that, the design of those two methods allows for probabilistic sensor fusion with other sensors, and the frequency of other sensors might vary significantly (e.g., camera at 10 Hz or GPS at 1 Hz). Thus, IMU based prediction might be conducted for different time periods in real applications.
In this test, we relied on the known IMU poses (position, rotation, and velocity), and used only IMU measurements to predict future poses. For both methods, we computed the RMSE for position, rotation, and velocity. 

The results are shown in Fig.~\ref{fig:prediction_error}, which demonstrate that our method achieves significantly improved accuracy compared to the competing one. This validates our theoretical analysis and our design motivation that a data driven method can potentially provide better IMU models for forward short-time integration compared to the hand-crafted method. The results also indicate that our method is more preferable for probabilistic sensor fusion.

\begin{figure}[t]
	\centering
	\includegraphics[width=.8\columnwidth]{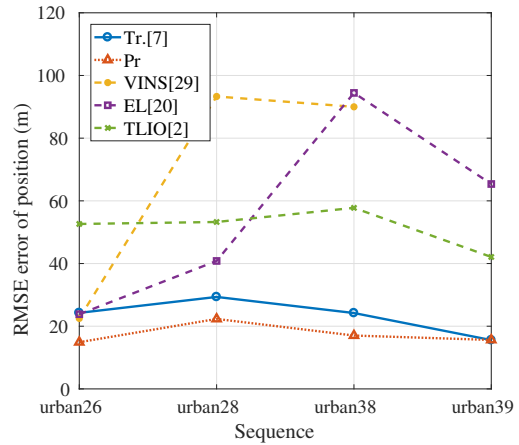}
	\caption{Positional RMSE in KAIST dataset (VINS~\cite{qin2018vins} fails on urban39).}
	\label{fig:vio_error}
\end{figure}

\begin{table}
	\scriptsize	
	\centering
	\begin{threeparttable}  
		\caption{\scriptsize{Average positional RMSE errors (m) for competing methods.}}  \label{table: vio_error}
		\begin{tabular}{l p{1.0cm}<{\centering}
				p{2.4cm}<{\centering}
				p{2.4cm}<{\centering}
			}  
			
			\toprule  
			\quad Dataset & KAIST & EuRoC & Ground robot \\
			
			\midrule
			\quad Tr.~\cite{farrell2008aided}  & 23.3482  & 0.1592 &  4.7007 \\
			\quad Pr. & \bf{17.4665}  & \bf{0.1210} &  \bf{3.7588} \\
			\quad VINS~\cite{qin2018vins} & 68.6 & 0.1778 & 8.2478 \\
			\quad EL~\cite{silva2019end} & 56.1167  & 2.0402 &  11.3905 \\
			\quad TLIO~\cite{liu2020tlio} & 51.4357 & 0.5262 & 9.8441 \\
			\bottomrule  
		\end{tabular}  
	\end{threeparttable}  
\end{table} 

\subsection{Tests on KAIST Dataset and Ground Robots}
To demonstrate the generality of our algorithm, we also conducted tests on KAIST dataset and ground robots.  
Fig.~\ref{fig:vio_error} shows positional RMSE per sequence in KAIST dataset and Table.~\ref{table: vio_error} shows average RMSE across different datasets. We emphasize that, for those testing cases, the proposed  algorithm  consistently  outperforms all competing methods by wide margins. Those results clearly show that our method is {\em not} application specific and can be applied in different use cases for performance gain.

\section{Conclusion}
To summarize, in this work we propose a framework with a data driven algorithm to obtain complex sensor model, which naturally outperforms hand-crafted ones and other deep learning based methods. In addition, our method regresses observable IMU kinematic terms, which is of theoretical guarantee and can be easily integrated for probabilistic sensor fusion. Both properties are not achieved by the competing DNNs. 
The experimental results on different testing platforms demonstrate that our algorithm is generally applicable and outperforms the competing state-of-the-art methods significantly.


\bibliographystyle{IEEEtran}
\bibliography{ref.bib}

\end{document}